\documentclass{article}
\usepackage{spconf,amsmath,graphicx}
\usepackage{amssymb}
\usepackage{graphicx,epsfig,setspace,subfig,url,amsmath}
\usepackage{algorithm}
\usepackage{algorithmic}
\usepackage{epstopdf}
\usepackage{amsmath}

\title{Deep Structured Learning for Mass Segmentation from Mammograms}
 \name{Neeraj Dhungel$^{\dagger}$ \qquad Gustavo Carneiro$^{\dagger}$ \qquad Andrew P. Bradley$^{\star}$ \sthanks{This work was partially supported by the Australian Research Council's Discovery Projects funding scheme (project DP140102794). Prof. Bradley is the recipient of an Australian Research Council Future Fellowship(FT110100623)}}
 \address {$^{\star}$ ACVT, School of Computer Science, The University of  Adelaide \\ $^{\dagger}$ School of Information Technology and Electrical Engineering, The University of Queensland }
\begin{document}

\ninept
\maketitle
\begin{abstract}
In this paper, we present a novel method for the segmentation of breast masses from mammograms exploring structured and deep learning.   
Specifically, using structured support vector machine (SSVM), we formulate a model that combines different types of potential functions, including one that classifies image regions using deep learning.    
Our main goal with this work is to show the accuracy and efficiency improvements that these relatively new techniques can provide for the segmentation of breast masses from mammograms.
We also propose an easily reproducible quantitative analysis to assess the performance of breast mass segmentation methodologies based on widely accepted accuracy and running time measurements on public datasets, which will facilitate further comparisons for this segmentation problem. 
In particular, we use two publicly available datasets (DDSM-BCRP and INbreast) and propose the computation of the running time taken for the methodology to produce a mass segmentation given an input image and the use of the Dice index to quantitatively measure the segmentation accuracy.
For both databases, we show that our proposed methodology produces competitive results in terms of accuracy and running time.

\end{abstract}
\begin{keywords}
Mammograms, mass segmentation, structured learning, structured inference
\end{keywords}

\vspace{-.2cm}
\section{Introduction}
\label{sec:intro}
\vspace{-.2cm}

Breast cancer is among the most common types of cancer in women. 
According to a WHO report~\cite{WHO}, breast cancer accounts for 22.9\% of diagnosed cancers and 13.7\% of cancer related death worldwide.  
Early detection of breast cancer using imaging techniques is vital to improve survival rates and the most commonly used screening technique 
is X-ray mammography (MG), which enables the detection of suspicious masses and micro-calcifications that are subsequently used for classification~\cite{Rahmati,Beller}.
It has been observed that there is a trade off between sensitivity and specificity in the manual analysis of MG, which in general can reduce the efficacy of the diagnosys process~\cite{elmore2009variability}.
The development of computer aided diagnostic (CAD) systems has the potential to improve this trade off, but it has been observed that the use of these systems in MG reduces the accuracy of screening by increasing the rate of biopsies without improving the detection of invasive breast cancer~\cite{Fenton_NEJM_09}. 
We believe that this issue can be fixed with a more easily reproducible and reliable assessment mechanism that provides a clear comparison between competing methodologies, which can lead to a better informed decision process related to the selection of appropriate algorithms for CAD systems in MG. 
Another reason for this poor performance lies in the reliance of current approaches on more traditional image processing and segmentation techniques, such as active contours, which typically produce sub-optimal results due to their non-convex cost functions and reliance on strong contour and appearance priors (e.g., smooth contours, strong edges, etc.). 
Therefore, we propose a statistical pattern recognition approach that estimates optimal (or near optimal) models directly from annotated data~\cite{carneiro_tmi_08}.

The main contribution of this paper is the use of structured support vector machine (SSVM) that can learn a structured output, representing the mass segmentation, from an input test image. We also propose a potential function (to be used in this structured learning problem) based on deep belief networks (DBN) that can learn complex features directly from MG.
We also propose an easily reproducible assessment that measures both the accuracy and the efficiency of breast mass segmentation methodologies on the publicly available databases INbreast~\cite{Inbreast} and DDSM-BCRP~\cite{DDSM1}. 
We show that our methodology produces competitive mass segmentation results of the field on these two databases.

\vspace{-.2cm}
\section{Methodology}
\vspace{-.2cm}

In this section, we describe our statistical model, SSVM for learning and the DBN-based potential function for mass segmentation.

\begin{figure}[t]
\begin{center}
\begin{tabular}{c}
\includegraphics[width=3in]{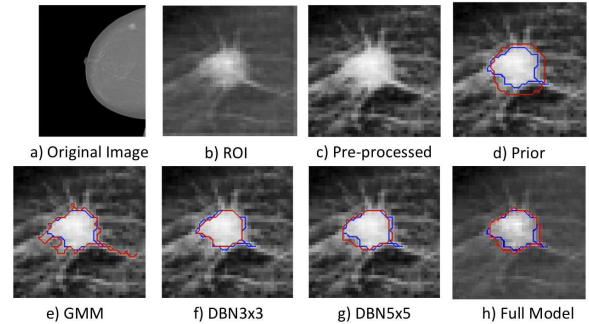} 
\end{tabular}
\end{center}

\caption{Examples of the results of each potential function and the final segmentation from the full model, where red denotes the automated segmentation and blue is the ground-truth contour. Note that $3 \times 3$ and $5 \times 5$ denote the input patch size (in pixels) for the DBN.}

\label{fig:potential_function_results}
\end{figure}

\vspace{-.2cm}
\subsection{Statistical Model for Mass Segmentation}
\vspace{-.2cm}

Let ${\cal X} = \{ {\bf x}_n \}_{n=1}^{N}$ be a collection of mammograms, with ${\bf x}:\Omega \rightarrow \mathbb R$ ($\Omega$ denotes the image lattice) representing the region of interest (ROI) in the MG containing the mass, and ${\cal Y} = \{ {\bf y}_n \}_{n=1}^{N}$ representing the segmentation of ${\bf x}_n$, with
${\bf y}:\Omega \rightarrow \{ -1,+1 \}$ ($+1$ represents mass and $-1$, background).  
Our model is denoted by the following probabilistic function~\cite{Szummer}:
\begin{equation}
P({\cal Y}|{\cal X},{\bf w}) = (1/Z)\exp\{ -\sum_n E({\bf y}_n , {\bf x}_n ; {\bf w} ) \},
\label{eq:CRF_model}
\end{equation}
where ${\bf w}$ represents the model parameters, and $Z$ the partition function.  
This model can be represented by a graph with ${\cal V}$ nodes and ${\cal E}$ edges between nodes, with $E(.)$ in (\ref{eq:CRF_model}) defined as:
\begin{equation}
\begin{split}
E({\bf y} , {\bf x} ; {\bf w} ) = \sum_{k=1}^K \sum_{i\in{\cal V}} w_{1,k}\phi^{(1,k)}( {\bf y}(i), {\bf x} ) \\+ 
\sum_{l=1}^L \sum_{i,j \in {\cal E}} w_{2,l}\phi^{(2,l)}({\bf y}(i),{\bf y}(j), {\bf x} ),
\label{eq:energy}
\end{split}
\end{equation}
where $\phi^{(1,k)}(.,.)$ represents one of the $K$ potential functions that links label (hidden) nodes and pixel (observed) nodes, $\phi^{(2,l)}(.,.,.)$ denotes one of the $L$ potential functions on the edges between label nodes, ${\bf w}=[w_{1,1},...,w_{1,K},w_{2,1},...,w_{2,L}]^{\top} \in \mathbb R^{K+L}$ and ${\bf y}(i)$ is the $i^{th}$ component of vector ${\bf y}$.  

\vspace{-.2cm}
\subsection{Structured Learning and Inference}
\vspace{-.2cm}

Learning the model parameters ${\bf w}$ in (\ref{eq:energy}) follows the SSVM procedure~\cite{Tsochantaridis_SSVM,Szummer}, as follows:
\begin{equation}
\begin{array}{l}
{\text{min. }} 
  \frac{1}{2}{||{\bf w}||}^2 +\frac{C}{N}\sum_n\xi_n \\
   {\text{s.t. }}E({\bf \hat{y}}_n, {\bf x}_n;{\bf w}) - E({\bf y}_n,{\bf x}_n ;{\bf w}) \geq\Delta({\bf y}_n,{\bf \hat{y}}_n) - \xi_n , \forall {\bf \hat{y}}_n \neq {\bf y}_n \\
   \;\;\;\;\;\; \xi_n \geq 0, \\
 \end{array}
\label{eq:SSVM_learning}
\end{equation}
where $\Delta(.,.)$ measures a distance in the label space, satisfying the conditions $\Delta({\bf y},{\bf y}_n) \geq 0$ and $\Delta({\bf y},{\bf y})= 0$.
This optimization is a quadratic programming problem involving an intractably large number of constraints.  In order to keep the number of constraints manageable, we use the cutting plane algorithm, where the most violated constraint for for the $n^{th}$ training sample is found by:
\begin{equation}
\hat{\bf y}_n = \arg \max_{{\bf y}} \Delta({\bf y}_n,{\bf y}) + E({\bf y},{\bf x}_n ;{\bf w})
\label{eq:constraint_generation}
\end{equation}
This algorithm is an iterative process that runs until no more violated inequalities are found (i.e., the right hand side in (\ref{eq:constraint_generation}) is strictly larger than zero).  This loss-augmented inference is efficiently solved with graph cuts~\cite{Boykov_graphcuts} if the function $\Delta(.,.)$ can be  decomposed in the label space.   A simple example that works with graph cuts is $\Delta({\bf y},{\bf y}_n) = \sum_i \delta( {\bf y}(i) - {\bf y}_n(i))$, which represents the Hamming distance that can be decomposed in the label space, with $\delta(.)$ denoting the Dirac delta function (this is the function used in this paper).

The label inference for {\bf x}, given the learned parameters {\bf w} from (\ref{eq:SSVM_learning}), is defined as follows:  
\begin{equation}
{\bf y}^*  = \arg\max_{\bf y} E({\bf y},{\bf x} ;{\bf w}),
\label{eq:inference}
\end{equation}
which can be efficiently solved for binary problems~\cite{Boykov_graphcuts}.

\vspace{-.2cm}
\subsection{Potential Functions}
\label{sec:potential_functions}
\vspace{-.2cm}

One of the advantages of learning the model parameter ${\bf w} $ in (\ref{eq:energy}) is that we can define and use any number of potential functions $\phi^{(1,k)}(.,.)$ between observed and hidden nodes and $\phi^{(2,l)}(.,.,.)$ between hidden nodes.  Specifically, we use three different types of potential functions between observed and hidden nodes.

The first type, $\phi^{(1,1)}(.,.)$, represents a prior of the location, size and shape of the mass (see Fig.~\ref{fig:potential_function_results}-(d)).  
This prior is the mean annotation estimated from the training set, as follows:
\begin{equation}
\phi^{(1,1)}({\bf y}(i),{\bf x}) = -\log P_p({\bf y}(i) = 1|\theta_p),
\label{eq:phi_1_1}
\end{equation}
where $P({\bf y}(i) = 1|\theta_p) = (1/N) \sum_n \delta({\bf y}_n(i) - 1)$.
The second potential function is represented by a generative model based on a Gaussian mixture model (GMM - see Fig.~\ref{fig:potential_function_results}-(e)), as in
\begin{equation}
\phi^{(1,2)}({\bf y}(i),{\bf x}) = -\log P_g({\bf y}(i) = 1|{\bf x}(i),\theta_g),
\label{eq:phi_1_2}
\end{equation}
where $P_g({\bf y}(i) = 1|{\bf x}(i),\theta_g) = (1/Z) \sum_{m=1}^M  \pi_m {\cal N}( {\bf x}(i) ; {\bf y}(i)=1, \mu_m, \sigma_m )P({\bf y}(i)=1)$ with $\theta_g$ denoting the parameters of the model (means $\mu_m$, variances $\sigma_m$ and weights $\pi_m$ of components), ${\cal N}(.)$ is the Gaussian function, $Z$ is the normalizer, ${\bf x}(i)$ represents the pixel value at image lattice position $i$, and $P({\bf y}(i)=1)=0.5$.
The model parameters $\theta_g$ in (\ref{eq:phi_1_2}) are learned from the annotated training set using the expectation-maximization (EM) algorithm~\cite{dempster1977maximum}.
Finally, the third function between observed and hidden nodes, $\phi^{(1,3)}(.,.)$, is based on the following free energy computed from a deep belief network  (DBN)~\cite{Hinton1}:
\begin{equation}
\phi^{(1,3)}({\bf y}(i),{\bf x}) = -\log P_d({\bf y}(i) = 1|{\bf x}_S(i),\theta_{d,S}),
\label{eq:phi_1_3}
\end{equation}
where ${\bf x}_S(i)$ represents a patch of size $S \times S$ pixels extracted around image lattice position $i$ (the reason for taking a patch from position $i$ instead of using the whole image is essentially to reduce the computational complexity of the training and inference procedures - see Fig.~\ref{fig:potential_function_results}(f)-(g), where (f) uses a region size $S=3$ and (g) uses $S=5$), $\theta_{d,S}$ represents the network weights and biases (hereafter, we drop the dependence on $\theta_{d,S}$ for notation simplicity).  The DBN is a generative model represented by a multi-layer perceptron containing a large number of layers (typically more than three) and a large number of nodes per layer. 
The underlying DBN model with $Q$ layers is represented by
\begin{equation}
\begin{array}{l}
P({\bf x}_S(i),  {\bf y}(i), {\bf h}_{1},...,{\bf h}_{Q}) = \\ 
\quad P({\bf h}_{Q}, {\bf h}_{Q-1}, {\bf y}(i))\left  ( \prod_{q=1}^{Q-2}P({\bf h}_{q+1}|{\bf h}_{q}) \right )P({\bf h}_{1}|{\bf x}_S(i)), \\ 
\label{eq:DBN}
\end{array}
\end{equation}
where ${\bf h}_q \in \mathbb R^{|q|}$ denotes the hidden variables at layer $q$ containing $|q|$ nodes.  The first term in (\ref{eq:DBN}) can be written as:
\begin{equation}
\begin{array}{l}
-\log( P({\bf h}_{Q},  {\bf h}_{Q-1}, {\bf y}(i) ) )\propto  \\ 
 \;\;\;\;\;\; - {\bf b}_{Q}^{\top}{\bf h}_{Q} - {\bf a}_{Q-1}^{\top}{\bf h}_{Q-1} - {\bf a}_{y}^{\top}[\frac{{\bf y}(i)+1}{2},\frac{1-{\bf y}(i)}{2}]^{\top} \\
 \;\;\;\;\;\; - {\bf h}_{Q}^{\top}{\bf W}{\bf h}_{Q-1}- {\bf h}_{Q}^{\top}{\bf W}_{y}[\frac{{\bf y}(i)+1}{2},\frac{1-{\bf y}(i)}{2}]^{\top}
\end{array}
\end{equation}
where ${\bf a}$, ${\bf b},{\bf W}$ are the biases and weights of the network, while the conditional probabilities in the remaining two terms can be factorized as 
$P({\bf h}_{q+1}|{\bf h}_{q}) = \prod_{i=1}^{|q+1|} P({\bf h}_{q+1} (i) |{\bf h}_{q})$ because the nodes in layer $q+1$ are independent from each other given ${\bf h}_{q}$, which is a consequence of the DBN structure (note that $P({\bf h}_{1}|{\bf x}_S(i))$ is similarly defined).  Finally, assuming that each node is activated by a sigmoid activation function $\sigma(.)$, we have
$P({\bf h}_{q+1}(i)|{\bf h}_{q}) = \sigma( {\bf b}_{q+1}(i) +  {\bf W}_i {\bf h}_{q} )$.  Then (\ref{eq:phi_1_3}) is computed with:
\begin{equation}
\begin{split}
P_d({\bf y}(i) = 1|{\bf x}_S(i)) \propto P_d({\bf y}(i) = 1, {\bf x}_S(i) ) =\\
 \sum_{{\bf h}_1} ... \sum_{{\bf h}_Q} P_d({\bf x}_S(i) , {\bf y}(i) = 1, {\bf h}_1,...,{\bf h}_Q),
\end{split}
\end{equation}
which can be estimated by the mean field approximation of the values in layers ${\bf h}_1$ to ${\bf h}_{Q-1}$ followed by the computation of free energy on the top layer~\cite{Hinton1}.
The training of the DBN involves the estimation of the parameter $\theta_{d,S}$ in (\ref{eq:phi_1_3}), which is achieved with an iterative layer by layer training of auto-encoders using contrastive divergence~\cite{Hinton1}.

We use two different types of potential functions between label (hidden) nodes in (\ref{eq:energy}), which encode label and contrast dependent labelling homogeneity.  More precisely, the first potential function in (\ref{eq:energy}) represents a label transition penalty~\cite{Szummer}, as follows:
\begin{equation}
\label{eq:phi_2_1}
\phi^{(2,1)}({\bf y}(i),{\bf y}(j),{\bf x})  = 1 - \delta ( {\bf y}(i) - {\bf y}(j)),
\end{equation}
and the second function denotes a contrast penalty, as follows:
\begin{equation}
\label{eq:phi_2_2}
\phi^{(2,2)}({\bf y}(i),{\bf y}(j),{\bf x})  = (1 - \delta ( {{\bf y}(i) - {\bf y}(j)} ))C({\bf x}(i), {\bf x}(j)),
\end{equation}
with ${\bf x}(i)$ representing the pixel value at position $i$, and $C({\bf x}(i), {\bf x}(j))=e^{-({\bf x}(i)-{\bf x}(j))^2}$.

Note that an interesting advantage about the model presented in (\ref{eq:CRF_model}) lies in its ability to accept and combine several different types of potential functions.  Fig.~\ref{fig:potential_function_results}(h) shows the final segmentation using all three potential functions.

\begin{algorithm}[t]
\caption{Training and Segmentation Algorithms}
\begin{algorithmic}
\label{alg:training}
\STATE {\bf Training Algorithm}
\STATE $\bullet$ Pre-process all images in ${\cal X}$ with the method described in \cite{ball2007digital}
\STATE $\bullet$ Learn parameters of potential functions $\phi^{(1,k)}(.,.)$ (for $k \in \{1,...,K\}$) using training sets ${\cal X}$ and ${\cal Y}$.
\STATE $\bullet$ Learn ${\bf w}$ in (\ref{eq:energy}) using SSVM optimization in (\ref{eq:SSVM_learning}).
\STATE {\bf Segmentation Algorithm}
\STATE $\bullet$ Given a test image ${\bf x}$, pre-process it~\cite{ball2007digital} and infer segmentation ${\bf y}^*$ with (\ref{eq:inference}) using graph cuts~\cite{Boykov_graphcuts}.
\end{algorithmic}
\end{algorithm}
\section{Materials and Methods}

The performance evaluation is carried out on two publicly available datasets: DDSM-BCRP~\cite{DDSM1} and INbreast~\cite{Inbreast}. 
The DDSM-BCRP~\cite{DDSM1} is part of the DDSM database used to evaluate CAD algorithms and consists of four datasets, from which we use the two datasets focused on spiculated masses, with 9 cases (77 annotated images) for training and 40 cases (81 annotated images) for testing. 
However, it is important to acknowledge that the annotations provided with DDSM-BCRP are inaccurate~\cite{horsch2011needs,Inbreast} and so most of the literature uses subsets of DDSM with bespoke annotations that are not publicly available. 
The recently proposed INbreast database~\cite{Inbreast} has been developed to provide a high quality publicly available mammogram database, containing accurate annotations. 
INbreast has a total of 56 cases containing 116 accurately annotated masses, which have been divided into mutually exclusive train and test sets, each containing 58 images each on training and testing set. It is important to note that some cases in DDSM-BCRP and INbreast database contain multiple masses, where each case presents the Craniocaudal (CC) and Mediolateral (MLO) views.

We use Dice index (DI) = $\frac{2TP}{FP+FN+2TP}$ for quantitatively measuring the segmentation accuracy. Here $TP$ denotes the number of mass pixels correctly segmented, $FP$ the background pixels falsely segmented as mass, $TN$ the correctly identified background pixels and $FN$ the mass pixels not identified. 
Finally, the running time reports the average execution time per image of the segmentation algorithm in Alg.~\ref{alg:training} on a standard computer (Intel(R) Core(TM) i5-2500k 3.30GHz CPU with 8GB RAM).
The ROI is produced by a manual annotation of the mass centre and scale, where the size of the final ROI is two times the manually annotated scale.  The ROI and is then resized to 40 x 40 pixels using bicubic interpolation. 
We adopt the image pre-processing described by Ball and Bruce~\cite{ball2007digital}. (see Fig.~\ref{fig:potential_function_results}(c)).  
This pre-processing step improves the contrast of the input image, which can potentially increase the separation between mass and background samples, facilitating the training and segmentation tasks.  

\vspace{-.2cm}
\section{Results}
\vspace{-.2cm}

\begin{figure}[t]
\begin{center}
\begin{tabular}{cc}
\includegraphics[width=3.5 in]{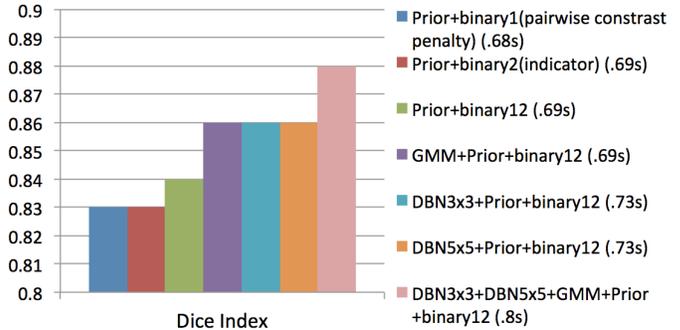} 
\end{tabular}
\end{center}
\vspace{-.2cm}
\caption{Dice index over the test set of INbreast and running time results of different versions of our model (in brackets).}
\vspace{-.2cm}
\label{fig:our results_alone}
\end{figure}

Figure~\ref{fig:our results_alone} shows the performance of several combinations of potential functions (Sec.~\ref{sec:potential_functions}) using the proposed model in (\ref{eq:CRF_model}) on INbreast~\footnote{Please note that our results on DDSM-BCRP are similar to the ones on the INbreast data.}. 
Specifically, Fig.~\ref{fig:our results_alone} shows that the best performance on the test set is obtained using a combination of all potential functions.  
Note that the Dice index of our methodology on the training set is $0.89$, which is similar to the performance on the test set shown in Fig.~\ref{fig:our results_alone}, which is $0.88$, indicating good generalization capability.
Also, the best Dice index of our methodology on the test set when we do not adopt the pre-processing described by Ball and Bruce~\cite{ball2007digital} is $0.85$, which indicates that this pre-processing is important in elevating the Dice index to $0.88$. 
Here, ``Prior", ``GMM", ``DBN" represent the functions $\phi^{(1,k)}$ for $k=\{1,2,3\}$ with $3 \times 3$ and $5 \times 5$ denoting the image patch size used by the DBN (see Equations~\ref{eq:phi_1_1}-\ref{eq:phi_1_3}), ``Binary1" is the label transition penalty in $\phi^{(2,1)}$ in~(\ref{eq:phi_2_1}), ``Binary2" the pairwise contrast penalty of $\phi^{(2,2)}$ in~(\ref{eq:phi_2_2}) and ``Binary12" indicates the use of both ``Binary1" and ``Binary2". 
Finally, the running time of each method is given in brackets.

Tab.~\ref{table:comparison} shows the accuracy and running time results of our approach with potential functions DBN3x3 + DBN5x5 + GMM + Prior + Binary12) on the test sets of DDSM-BCRP and INbreast. 
The results from the other methods are as reported by Horsh et al.\cite{horsch2011needs} or by their original authors.  
However, note that the majority of the results on DDSM cannot be compared directly because they have been obtained with train and test sets that are not publicly available, and so cannot be reproduced (indicated by ``Reproducible"). Also not all performance measures were reported (indicated by ``?").

\begin{table}[t]
\caption{Comparison between the proposed and several state-of-the-art methods.} 
\centering
\scriptsize
\begin{tabular}{|c| c| c| c| c| c|}
\hline 
Method & Rep. & Images & Dataset & DI & Time\\
\hline
\quad Proposed & yes&158&DDSM-BCRP&0.87&0.8s\\
\hline
\quad Beller et al.~\cite{Beller} & yes & 158&DDSM-BCRP& 0.70 & ?? \\
\hline
\quad  Ball et al.~\cite{ball2007digital} & no & 60&DDSM  & 0.85 & ? \\
\hline
\quad Hao et al.~\cite{Hao}& no & 1095&DDSM &0.85 &5.1s \\
\hline
\quad Rahmati et al.~\cite{Rahmati}& no & 100 & DDSM & 0.93 & ? \\
\hline
\quad Song et al.~\cite{Song} & no  & 337 & DDSM & 0.83 & 0.96s \\ 
\hline 
\quad Yuan et al.~\cite{Yuan} & no &  483 & DDSM &  0.78& 4.7s  \\
\hline 
\quad Proposed&yes&116&INbreast&0.88&0.8s\\ 
\hline
\quad Cardoso et al.~\cite{Cardoso}&yes&116&INbreast & 0.88 & ? \\ 
\hline 
\end{tabular}
\label{table:comparison}
\end{table}

\vspace{-.2cm}
\section{Discussion and Conclusion}
\vspace{-.2cm}

Fig.~\ref{fig:our results_alone} demonstrates that segmentation accuracy improves with the introduction of each potential function at a relatively small
computational cost.  Also, our method shows good generalization ability given the small differences between results on the train and test sets.  
It is interesting to note that the pre-processing stage provides a substantial increase in accuracy.
Moreover, the ``Prior" and ``DBN" potential functions produce the best results among the potential functions $\phi^{(1,k)}$ in Sec.~\ref{sec:potential_functions}, but their integration in the  model (\ref{eq:CRF_model}) is essential to produce the state-of-the-art results displayed in Fig.~\ref{fig:our results_alone}. Also notice that although the Dice index showed by the ``Prior" potential function is relatively high, the shape produced by ``Prior" is mostly circular, which means that DBN and GMM play important roles in segmenting the irregular boundaries shown by breast masses.

The comparison with the state of the art shown in Tab.~\ref{table:comparison} shows that our approach is computationally efficient, running in 0.8 seconds.  In fact, this is the most efficient methodology reported in the field for this problem, to the best of our knowledge.
Our method shows the best results on DDSM-BCRP, but using other subsets and annotations from DDSM (that are not publicly available), our method still appears competitive, having the second best overall result, with~\cite{Rahmati} being the most accurate.  However, because we do not have access to the annotations and images used in~\cite{Rahmati}, it is impossible to reproduce their experiment, making a direct comparison difficult. 
Finally, on INbreast our method ties with the approach by Cardoso et al ~\cite{Cardoso}, which is the current state of the art.
The main limitation affecting our algorithm on both databases is the small size of the training set and the limited appearance and shape variations of the mass in this training set. 
These two issues induce the learning algorithm to put more weight on the potential function encoding the shape prior, $\phi^{(1,1)}(.,.)$, in (\ref{eq:energy}), which results in large bias and small variance.  By increasing the training sets, we can reduce the bias significantly without necessarily increasing the variance.  This aspect is worth noticing because it increases the potential of our approach to produce more accurate results if richer and larger training sets become available. 

We have shown that structured and deep learning produces competitive results on breast mass segmentation in terms of accuracy and efficiency.  
We strongly recommend that other researchers interested in the problem of breast mass segmentation use of the publicly available annotated databases DDSM-BCRP and 
INbreast.  This will allow clearer comparisons between different methodologies, which can be used in determining the most effective approaches for this problem.

\bibliographystyle{IEEEbib}
\bibliography{ISBI_Mass}

\end{document}